# EVALUATION OF UNCERTAIN INFERENCE MODELS I: PROSPECTOR*

Robert M. Yadrick
Bruce M. Perrin
David S. Vaughan
McDonnell Douglas Astronautics Company
P.O. Box 516, St. Louis, MO 63166

Peter D. Holden
Karl G. Kempf**
McDonnell Douglas Research Laboratories
P.O. Box 516, St. Louis, MO 63166

## ABSTRACT
This paper examines the accuracy of the PROSPECTOR model for uncertain reasoning. PROSPECTOR's solutions for a large number of computer-generated inference networks were compared to those obtained from probability theory and minimum cross-entropy calculations. PROSPECTOR's answers were generally accurate for a restricted subset of problems that are consistent with its assumptions. However, even within this subset, we identified conditions under which PROSPECTOR's performance deteriorates.

## INTRODUCTION

Researchers in artificial intelligence have proposed or implemented several approaches to uncertain reasoning for knowledge-based systems. MYCIN [1], PROSPECTOR [2], EMYCIN [3], and AL/X [4] combine evidence and propagate beliefs by using heuristic indices and making admittedly questionable assumptions. Other approaches involve adaptations of fuzzy set theory [5], Dempster-Shafer belief functions [6], and set-covering theory [7]. Still others include INFERNO [8] and endorsement theory [9]. Unfortunately, there is no consensus on which approach is best (or even suitable) for any particular application. One reason may be that virtually no rigorous empirical evidence exists concerning their accuracy under even ideal conditions.

There is a long history of research to evaluate the performance of inferential statistics under a variety of conditions that may be encountered in real-world applications. One important approach uses artificial or simulated data in which known parameters are varied in systematic ways so that correct outcomes can be calculated [e.g., 10]. Evaluation studies of this sort yield valuable insights into the relative strengths and weaknesses of the statistics under study. The ad hoc uncertainty indices of knowledge-based systems are analogous to inferential statistics -- they reflect probabilistic states of a condition. The adequacy of such indices can be assessed by studying their response to changes in simulated data.

This paper is the first in a series that will examine current uncertainty models by adopting a comparable rationale and using methods similar to those found in statistical evaluation studies. Our analyses are based upon the study of a large number of very simple inference networks that consist of two pieces of evidence and one conclusion. Such networks constitute one of the basic building blocks of larger networks, but are small enough to allow for detailed explication of the sources of error. Evaluation of these networks requires both propagation and combining functions. Error is therefore symptomatic of problems that can accrue when many pieces of evidence bear on a conclusion or effects are propagated through several links in an inference chain. We will discuss the issues involved in extending our analyses to larger networks later in this paper.

We focus on the fundamental accuracy of the PROSPECTOR model. By basing PROSPECTOR indices on known probability values, we eliminated a major source of error in an actual application -- human estimation of probabilities. We then compared PROSPECTOR's solutions to the statistically correct solutions produced by a minimum cross-entropy inference procedure [11]. Moreover, by examining a large number of networks, we are conceptually evaluating PROSPECTOR's ability to deal with a population of problems, i.e., the reliability of a PROSPECTOR-based system in operational use.

## OVERVIEW OF THE PROSPECTOR MODEL

A brief explanation of PROSPECTOR's model is in order before our methods and results are discussed. We will consider only the essential aspects of PROSPECTOR that deal with issues of combining evidence and propagating the effects of new evidence throughout the network. A number of features that are not directly related to evidence propagation or combination but that may nonetheless affect accuracy (e.g., calculations performed upon user responses to system inquiries), are not addressed here. Additionally, only the formulas used to handle uncertain evidence will be presented; the equations for certain evidence are simplifications of these formulas.

The basic formula PROSPECTOR uses to compute the conditional probability of a conclusion given new evidence is as follows:

$$P'(C|E) = \begin{cases} P(C|\bar{E}) + \dfrac{P(C) - P(C|\bar{E})}{P(E)} \times P'(E) & \text{FOR } 0 \leq P'(E) < P(E) \\ \\ P(C) + \dfrac{P(C|E) - P(C)}{1 - P(E)} \times [P'(E) - P(E)] & \text{FOR } P(E) \leq P'(E) \leq 1 \end{cases}$$

In this equation, $P'(C|E)$ is the conditional probability of the conclusion that is inferred given new information; $P(C|E)$ is the original conditional probability of the conclusion, given that the evidence is certainly true; $P(C|\bar{E})$ is the original conditional probability of of the conclusion, given that the evidence is certainly false; $P(C)$ is the base rate (prior probability) of the conclusion; $P(E)$ is the base rate of the evidence; and $P'(E)$ is the new probability (calculated from a user response to a request for diagnostic information) of the evidence. While this formula is written in terms of evidence bearing on a conclusion, the conclusion or the evidence could just as easily be an intermediate hypothesis in an inference chain. The equation essentially defines $P'(C|E)$ as a piecewise linear function anchored at 0, $P(E)$, and 1. Intermediate values are interpolated.

The new overall probability for evidence $P'(E)$ is calculated in one of three ways when more than one piece of evidence bears on the conclusion. How it is

---

calculated depends on the hypothesized relationship between the pieces of evidence and the conclusion. First, if the conclusion follows only if all pieces of evidence are believed true to some degree, then a conjunctive ("AND") rule is applied. In this case, P'(E) = MIN [(P'(Ei)], where the Ei are the various pieces of evidence. Second, if the conclusion follows if any of the pieces of evidence are true, a disjunctive ("OR") rule is applied and P'(E) = MAX [P'(Ei)]. In either case, P'(E) is used in the basic equation to estimate P'(C|E).

The third rule for determining the new evidence probability is to assume that each piece of evidence has an independent effect upon the conclusion. In this case, PROSPECTOR uses each P'(Ei) separately, yielding a set of P'(C|Ei). These conditional probabilities are then converted to odds according to the formula:
$$O'(C|Ei) = P'(C|Ei) / [1 - P'(C|Ei)].$$
These odds are converted to "effective likelihood ratios" by this formula:
$$L'i = O'(C|Ei) / O(C),$$
where O(C) is the odds of the conclusion. Next, the individual effective likelihood ratios are combined using this heuristic equation:
$$O'(C|E) = (\prod L'i) O(C).$$
Finally, this odds is converted to a probability by the formula:
$$P'(C|E) = O'(C|E) / [1 + O'(C|E)],$$
yielding PROSPECTOR's estimate of the conditional probability of the conclusion given two or more pieces of independent evidence.

## METHOD

An inference network can be represented as a multi-dimensional contingency table [12] which has a dimension for each piece of evidence and each conclusion. Each cell in the table contains the joint probability for the associated states of pieces of evidence and conclusions.

In real-world applications, it may be difficult and sometimes impossible to obtain satisfactory estimates of these cell entries. This was one of the motivations for the development of pseudo-Bayesian models like PROSPECTOR and MYCIN. (Several researchers have since developed methods for overcoming these problems in most cases, e.g., [13], [14].)

For theoretical studies, however, there is no difficulty in generating simulated data for such tables. If these tables are taken as representing actual situations, one may focus directly on the question of how well a particular inference model can approximate correct answers. Tables that exhibit a variety of potentially interesting properties can be produced. For example, the degree of association (i.e., conditional dependence) between pieces of evidence and conclusions can be examined systematically.

Figure 1 shows a contingency table which represents a two-evidence, one conclusion network. We implemented a problem generator to produce sets of small networks in contingency table form. The process for producing contingency tables was necessarily somewhat different for tables representing associated and independent evidence nodes.

For associated evidence nodes, the base rate for each piece of evidence and the conclusion was set randomly to a number between zero and one. Then each cell entry was likewise set randomly between zero and one. The table was rescaled by an algorithm called iterative proportional fitting [12]. The base rates and associations in the resulting tables were both randomly assigned and independent of each other. This assured that any error effects resulting from one factor (e.g., conclusion base rate) could not be attributed to any other table characteristic. It also assured that a full range of base rates and associations would be sampled.

| EVIDENCE 1 | EVIDENCE 2 | CONCLUSION FALSE | CONCLUSION TRUE | MARGINALS |
|---|---|---|---|---|
| FALSE | FALSE | P($\bar{E_1}$ & $\bar{E_2}$ & $\bar{C}$) | P($\bar{E_1}$ & $\bar{E_2}$ & C) | P($\bar{E_1}$ & $\bar{E_2}$) |
| FALSE | TRUE | P($\bar{E_1}$ & $E_2$ & $\bar{C}$) | P($\bar{E_1}$ & $E_2$ & C) | P($\bar{E_1}$ & $E_2$) |
| TRUE | FALSE | P($E_1$ & $\bar{E_2}$ & $\bar{C}$) | P($E_1$ & $\bar{E_2}$ & C) | P($E_1$ & $\bar{E_2}$) |
| TRUE | TRUE | P($E_1$ & $E_2$ & $\bar{C}$) | P($E_1$ & $E_2$ & C) | P($E_1$ & $E_2$) |

Figure 1. Contingency Table for a 3-Node Network

For independent evidence nodes, the table marginals must equal the product of the base rates (i.e., the joint probability is the product of the simple probabilities.). The first step was to compute marginals from the base rates, and then to randomly apportion each marginal between the two corresponding table cells. For example, the P($\bar{E1}$ & $\bar{E2}$) marginal in Figure 1 was apportioned between the two first-row cells. The resulting table exhibited both statistical independence between pieces of evidence and random associations between each piece of evidence and the conclusion. We generated four hundred independent and four hundred associated networks using these procedures.

Initial analysis showed that many networks contained counterintuitive relationships, e.g., indicating support of the conclusion if one piece of evidence was true but negating the conclusion if both were true. We also found that PROSPECTOR's error for such networks often exceeded .50. PROSPECTOR apparently does not model these situations adequately (Shortliffe and Buchanan [1] noted explicitly that MYCIN was not capable of modeling such phenomena). To provide a conservative test, these problems must be considered outside of PROSPECTOR's domain. Therefore, we required that the networks exhibit one of the following patterns of conditional probabilities:
P(C|$\bar{E1}$ & $\bar{E2}$) ≥ P(C|$\bar{E1}$ & E2), P(C|E1 & $\bar{E2}$) ≥ P(C|E1 & E2)   [1]
or
P(C|$\bar{E1}$ & $\bar{E2}$) ≤ P(C|$\bar{E1}$ & E2), P(C|E1 & $\bar{E2}$) ≤ P(C|E1 & E2)   [2]
These restrictions left 66 independent and 73 associated networks.

Each remaining network was solved for a set of "new evidence probabilities." PROSPECTOR updates probabilities when users respond to system inquiries. Some probabilities change as a direct result of responses, while others change as a result of propagation. For present purposes, we simply assigned new probabilities to each evidence node. These nodes independently assumed values of 0.0, 0.2, 0.5, 0.8, and 1.0, in turn, to simulate the wide range of probabilities that could arise as a result of user responses. Thus each network was solved for 25 sets of new evidence probabilities. Our analysis is based upon a total of 3,475 test cases.

Each test case was solved twice: once by a program that implemented the PROSPECTOR model described in section II (and gave answers for each of the three rule sets), and once by an inference procedure we developed for this comparison. The details of this inference engine are discussed elsewhere [11]. Briefly, the joint probabilities in the original table are updated in a manner which preserves the original table's patterns of association. The resulting, updated table is comprised of the minimum cross-entropy transformations of the original entries, given the new evidence [15]. They are the statistically correct answers, under the principles of entropy theory [16].

## RESULTS AND DISCUSSION

It was intended by PROSPECTOR's designers that answers be reasonably close approximations to those that would result from a rigorous probability analysis, if one could be performed. Consequently, we have focused on measuring the average PROSPECTOR error for each network over the set of new evidence updates. We defined error as the absolute difference between each correct answer and the corresponding PROSPECTOR estimate. Finally, we also examined the maximum error, i.e., the greatest error resulting from a single new evidence probability update.

A case study will illustrate these points. Suppose that a network is generated for which the base rate of the conclusion and the base rates of each piece of evidence are all equal, say $P(C) = P(E1) = P(E2) = 0.50$. Further, the network was produced by the procedure that yields independent evidence. Also, the truth of either piece of evidence alone tends neither to strongly support nor negate the conclusion. Finally, the conclusion is rather strongly suggested if both pieces of evidence are true. All this is so if, say, $P(C|\overline{E1} \& \overline{E2}) = .10$; $P(C|\overline{E1} \& E2) = P(C|E1 \& \overline{E2}) = .50$; and $P(C|E1 \& E2) = .90$. Figure 2 shows the contingency table representation of this problem (Part A) and plots PROSPECTOR's error (correct - PROSPECTOR) across the range of new E1 and E2 probabilities (Part B).

This particular example is rather unrealistic (e.g., all base rates of 0.50), but does illustrate a few pertinent points. First, the average signed error over all new probability values is zero, showing that averaging over signed errors can be particularly misleading. Second, the average absolute difference (unsigned error) is approximately .0098. This can be considered the expected error for this network if all new evidence probabilities (i.e., user responses) are thought to be equally likely over the life of the system. The actual error PROSPECTOR would exhibit in operational use, of course, depends on how the new evidence probabilities are distributed. Finally, the maximum error for this network is about .055. Because this table and the error plot are symmetrical, the maximum error occurs twice. This will not generally be the case. For this example, however, the maximum error occurs when the user responds that both pieces of evidence are either definitely true or definitely false. It is conceivable that the distribution of user responses in many cases would be skewed toward these ends of the scale. That is, users could be expected to respond that evidence is present or absent more often than they might respond that they are uncertain. If so, PROSPECTOR's answers for this particular network would more frequently approach the maximum error than the average error.

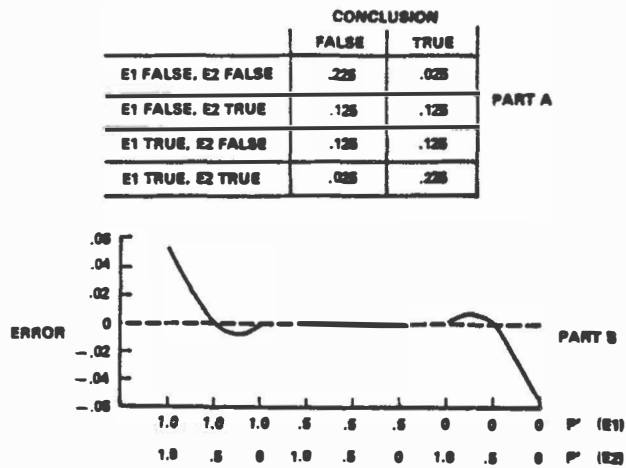

Figure 2. PROSPECTOR Case Study 1

### 1. PROSPECTOR Rule Sets.

The overwhelming majority of our sample problems were solved most accurately by PROSPECTOR's independence rule. The actual independence (or lack of it) inherent in the data does little to determine which rule set works best in terms of reducing overall error. As Table 1 shows, 61 of 66 independent evidence examples and 58 of 73 associated evidence examples were solved best using independence rules. The errors shown are the averages, over all problems, of the two error measures discussed above.

TABLE 1. MOST ACCURATE PROSPECTOR RULE
SET FOR INDEPENDENT AND ASSOCIATED EVIDENCE

| Relation of Evidence | Rule Set Type | | | Overall Average Error | Overall Maximum Error |
|---|---|---|---|---|---|
| | Conjunctive | Disjunctive | Independent | | |
| Independent | 5 | 0 | 61 | .014 | .055 |
| Associated | 9 | 6 | 58 | .022 | .083 |

Average absolute error is low. It is particularly low when the evidence is independent, rather than associated. In any event, PROSPECTOR's estimates are quite accurate most of the time. However, maximum error averages are considerably higher. This suggests that some user responses result in relatively inaccurate solutions, a point that will be examined in greater detail later in this paper. For the present, we turn to a discussion of what factors result in each of PROSPECTOR's rule sets being relatively accurate or inaccurate.

### 2. Error in Conjunctive and Disjunctive Rule Sets.

The adequacy of PROSPECTOR's conjunctive rule was found to rest heavily upon the degree to which:
$P(C|\overline{E1} \& \overline{E2}) \approx P(C|\overline{E1} \& E2) \approx P(C|E1 \& \overline{E2})$.
This is so because PROSPECTOR approximates each of these separate conditional probabilities by a single formula:

$$\frac{P(\overline{E1} \& \overline{E2} \& C) + P(\overline{E1} \& E2 \& C) + P(E1 \& \overline{E2} \& C)}{1 - [P(E1) * P(E2)]}$$

The disjunctive rule set generally requires networks for which:
$P(C|\overline{E1} \& E2) \approx P(C|E1 \& \overline{E2}) \approx P(C|E1 \& E2)$.
PROSPECTOR approximates each of these conditional probabilities by the formula:

$$\frac{P(\overline{E1} \& E2 \& C) + P(E1 \& \overline{E2} \& C) + P(E1 \& E2 \& C)}{1 - [P(\overline{E1}) * P(\overline{E2})]}$$

By examining our data, we found that the respective conditional probabilities for a given problem must conform closely to these ideals. Even relatively small

335

variations from equality between these conditional probabilities result in the independence rule set being more accurate than either the conjunctive or disjunctive rule sets.

The size of conjunctive or disjunctive error depends in large measure upon two factors, each of which interacts with the other to mitigate or increase the impact of the other. The first factor is the degree to which the actual conditional probabilities given above are approximated by the corresponding PROSPECTOR formulas. This condition will be met only when the conditional probabilities are equal and the evidence is independent.

The second factor is the difference between the fourth conditional probability and the average of the other three. For example, with a disjunctive rule set the difference between $P(C|\overline{E1} \& \overline{E2})$ and the average of $P(C|\overline{E1} \& E2)$, $P(C|E1 \& \overline{E2})$, and $P(C|E1 \& E2)$ is critical. Larger inaccuracies result as this difference becomes larger, unless the first condition is fully satisfied.

3. Independence Rule Sets.

It is considerably more difficult to identify the sources of error for networks solved by the independence rule set. It is possible, however, to write an equation for error in such cases. We will use a simplified notation here, since the expressions become very long otherwise. Let:
   $Pi$ = joint probabilities, e.g., $P1 = P(\overline{E1} \& \overline{E2} \& \overline{C})$, as shown in Figure 1.
   $Bi$ = evidence base rates, e.g., $B1 = P(E1)$
   $\overline{Bi}$ = 1 - $Bi$
   $Ci$ = new evidence probabilities, e.g., $C1 = P'(E1)$
   $\overline{Ci}$ = 1 - $Ci$

Even with this notation, PROSPECTOR's independence rule solution for our 3-node networks is long, and is given in Figure 3. The correct answer is
$[(P2 / \overline{B1} * \overline{B2}) * \overline{C1} * \overline{C2}] + [(P4 / B1 * \overline{B2}) * C1 * \overline{C2}] + [(P6 / \overline{B1} * B2) * \overline{C1} * C2] + [(P8 / B1 * B2) * C1 * C2]$,
if the evidence is independent. Note that the independence rule set often provided the best solution even when evidence was not independent (Table 1). This formula, however, cannot be used to compute the correct answer for associated evidence cases because joint probabilities cannot be obtained by multiplying simple probabilities. Unfortunately, it is not easy to look at the formula and quickly estimate which network configurations will produce sizeable error and which will not.

$P(C) = \{[(P2+P4)(B1\overline{C1}) + (P6+P8)(\overline{B1} C1)] \times [(P2+P6)(B2 \overline{C2}) + (P4+P8)(\overline{B2} C2)] \times \overline{B3}\}$

$+ \left(\{[(P2+P4)(B1 \overline{C1}) + (P6+P8)(\overline{B1} C1)] \times [(P2+P6)(B2 \overline{C2}) + (P4+P8)(\overline{B2} C2)] \times \overline{B3}\}\right.$

$+ \{[(B1 \overline{B1}) - (P2+P4)(B1 \overline{C1}) - (P6 + P8)(\overline{B1} C1)]$

$\left. \times [(B2 \overline{B2}) - (P2+P6)(B2 \overline{C2}) - (P4+P8)(\overline{B2} C2)] \times B3\}\right)$

Figure 3. PROSPECTOR Independence Rule Solution

It is possible to identify a single factor which relates strongly to independence rule set error. PROSPECTOR's estimates are increasingly inaccurate as the association between evidence and conclusion becomes stronger. PROSPECTOR is most accurate in trivial cases in which uncertain inference is unnecessary. Stated another way, error is smallest when the conditional probability of the conclusion is approximately the same whether the evidence is true or not. This relationship is shown in Figure 4, which plots a function we fit to our error data.

The function shows the relationship between a simple measure of the strength of the evidence-conclusion association (ABS $[P(C|\overline{E1} \& \overline{E2}) - P(C|E1 \& E2)]$) and error. Thus, the function averages across a wide range of values for the base rates, new evidence probabilities, and conditional probabilities.

To illustrate the significance of the function in Figure 4, we developed a fairly realistic case study. Suppose that a three-node network must represent the situation in which all evidence and conclusion base rates are low. Further, the conclusion is likely true if either piece of evidence is true and very likely true if both pieces of evidence are true. Finally, the two pieces of evidence are independent of each other. This situation approximates several sub-networks in the examples of real mineral exploration networks given in a report on the PROSPECTOR project [2].

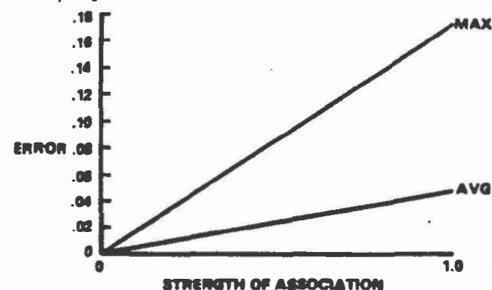

Figure 4. Error as a Function of Associative Strength

For example, let $P(E1) = .01$, $P(E2) = .02$, $P(C) = .05$, $P(C|E1) = .60$, $P(C|E2) = .70$, and $P(C|E1 \& E2) = .95$. Figure 5 shows the only contingency table that represents this situation exactly.

|  | CONCLUSION | |
| --- | --- | --- |
|  | FALSE | TRUE |
| E1 FALSE, E2 FALSE | .94581 | .00819 |
| E1 FALSE, E2 TRUE | .00589 | .01391 |
| E1 TRUE, E2 FALSE | .00389 | .00871 |
| E1 TRUE, E2 TRUE | .00081 | .00919 |

Figure 5. Case Study 2 -- Contingency Table

Rather than summarizing unsigned error in this situation, we present Figure 6, which plots PROSPECTOR's error across the range of new evidence probabilities $P'(E1)$ and $P'(E2)$. It is apparent that PROSPECTOR's answers for a considerable range of new probabilities are very inaccurate. Again, some of this error results from complex interactions between conditional probabilities, base rates, new probabilities, etc. But the most important determinant of the level of error is simply the large difference between $P(C|\bar{E1}\ \&\ \bar{E2})$ and $P(C|E1\ \&\ E2)$, a rough measure of the association between the evidence and the conclusion.

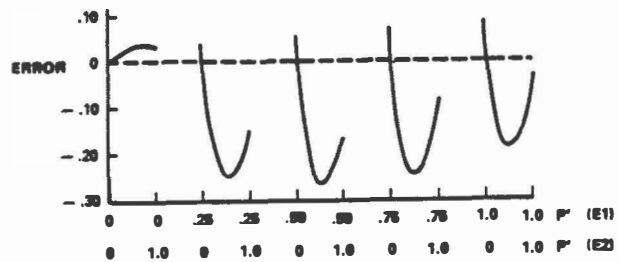

Figure 6. Case Study 2 -- Error Plot

A logical extension to this study would be to examine PROSPECTOR's accuracy with networks involving several pieces of evidence, intermediate nodes, and conclusions. However, the issues involved in such extensions are deceptive in their apparent simplicity. The important matter for bigger networks is simply the final amount of error, which reflects the degree to which errors are compounded or canceled as effects are propagated through the network and new probabilities are assigned to intermediate nodes.

It is hard to predict in advance just what insights might result from this effort. It may be difficult to discover general principles, and findings would be of little interest if causes of error can be determined only on a case-by-case basis. Additionally, we could deliberately configure networks to yield large or small amounts of error. This would not be very informative for anyone interested in a specific application.

Another matter concerns just how to represent particular networks so as to constitute a rigorous yet fair test for PROSPECTOR. A given contingency table can be interpreted using various combinations of PROSPECTOR conjunctive, disjunctive, and independence rules. The number of possible combinations could quickly become unmanageable as the size of the network increases.

## SUMMARY

Several summary conclusions can be drawn from our analysis of PROSPECTOR. First, in fairness, PROSPECTOR is satisfactorily accurate in many instances within its problem domain (i.e., consistent with the restrictions given by ineqalities 1 and 2 above). Even so, another important conclusion is that the networks that are least-well represented by PROSPECTOR are those in which evidence strongly influences the probablility of the conclusion. It seems reasonable that these will be the cases of critical interest in implementing a PROSPECTOR-based system. Furthermore, for any given network, new evidence probabilities can either compound or mitigate this problem. This means that accuracy in practice may be undeterminable if the system builder does not roughly know the distribution of expected user responses.

Independence rules generally provide the best PROSPECTOR solutions. It would be imprudent to suggest that conjunctive and disjunctive rules not be used in practice. But it is not unreasonable to suggest that networks be examined, if possible, to determine the degree to which the appropriate conditional probabilities are equal before such rules are incorporated.

Finally, it is doubtful that a single rule set could be found that would adequately handle different sets of new evidence probabilities for a given network. In practice, this means that a system developed using, for example, cases in which a piece of evidence is predominantly true would work poorly on cases in which that evidence is false. If cases with both positive and negative evidence were used in system development, it could be impossible to identify a single rule set that would work consistently. These problems could be very difficult to resolve.

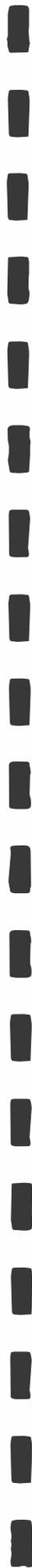